\documentclass[%
 reprint,
 superscriptaddress,
 amsmath,amssymb,
 aps,
longbibliography,
]{revtex4-2}

\usepackage{graphicx}
\usepackage{dcolumn}
\usepackage{bm}
\usepackage{natbib}
\setcitestyle{numbers,sort&compress}
\begin{document}

\preprint{APS/123-456}

\title{Performance-guaranteed regularization in maximum likelihood method: Gauge symmetry in Kullback -- Leibler divergence}

\author{Akihisa Ichiki}
\email{ichiki@fukuoka-u.ac.jp}
\affiliation{Department of Applied Mathematics, Faculty of Science, Fukuoka University, 8-19-1, Nanakuma, Jonan-ku, Fukuoka City, 814-0180, Japan}
\date{\today}

\begin{abstract}
The maximum likelihood method is the best-known method for estimating the probabilities behind the data. However, the conventional method obtains the probability model closest to the empirical distribution, resulting in overfitting. Then regularization methods prevent the model from being excessively close to the wrong probability, but little is known systematically about their performance. The idea of regularization is similar to error-correcting codes, which obtain optimal decoding by mixing suboptimal solutions with an incorrectly received code. The optimal decoding in error-correcting codes is achieved based on gauge symmetry. We propose a theoretically guaranteed regularization in the maximum likelihood method by focusing on a gauge symmetry in Kullback -- Leibler divergence. In our approach, we obtain the optimal model without the need to search for hyperparameters frequently appearing in regularization. 
\end{abstract}

\maketitle

\section{Introduction}

The maximum likelihood method is often employed to estimate the ground truth distribution $P_{\rm GT}(x)$ ($x\in\mathbb{R}^D$) from given data that generates them. The basic criterion is to find a probability model $P(x)$ that minimizes the Kullback -- Leibler divergence~(KL divergence) between the probability model $P$ and the underlying distribution $P_{\rm GT}$~\cite{kullback1951information, bishop2006pattern}: 
\begin{eqnarray}
D[P_{\rm GT}, P] := \int \,dx P_{\rm GT}(x)\ln\dfrac{P_{\rm GT}(x)}{P(x)}\,.
\end{eqnarray}
Since the ground truth $P_{\rm GT}$ is unknown and to be estimated, the well-known maximum likelihood method employs the empirical distribution $P_{\rm emp}^\xi$ as an approximation of $P_{\rm GT}$. However, the probability model $P$ that minimizes the KL divergence $D[P_{\rm emp}^\xi, P]$ results in the model far from $P_{\rm GT}$ because of the difference between the empirical distribution $P_{\rm emp}^\xi$ and the ground truth $P_{\rm GT}$. This phenomenon is well-known in machine learning as overfitting~\cite{hinton1993keeping, JMLR:v15:srivastava14a}. To avoid overfitting, regularization is applied to prevent $D[P_{\rm emp}^\xi, P]$ from becoming too small~\cite{tibshirani1996regression, hoerl1981ridge, zou2005regularization, zhang2010nearly}. Note that much of the success of regularization is backed by empirical evidence, and little is guaranteed theoretically on its performance. For the maximum likelihood estimation as model selection, Akaike's information criterion~(AIC) gives a theoretically guaranteed regularization~\cite{akaike1974new, burnham2011aic, konishi2008information}. The method proposed in the present paper shares the same root with AIC. While AIC is defined based on an unbiased estimator of the KL divergence between the model and the ground truth, our method does not require such a quantity explicitly.

 Regularization to prevent overfitting is similar to error-correcting codes~\cite{hamming1950error, reed1960polynomial, gallager1968information}. In error-correcting codes, a sender encodes the message redundantly, and a receiver decodes it by finding the solution to an optimization problem corresponding to the received code. However, since the communication channel is noisy, the received code contains errors. Thus the optimization problem corresponding to the received code does not give correct decoding. The receiver utilizes a mixture of non-optimal solutions to the optimization problem corresponding to the received code to decode the correct message~\cite{chung2001design, bayati2011dynamics}. In correspondence with physics, the optimization problem corresponds to the search for the ground state for the Hamiltonian, which the received code corresponds to. Since the Hamiltonian corresponding to the received code is incorrect, the correct message corresponds to the excited state for the incorrect Hamiltonian. In this situation, a mixture of excited states at finite temperature is more appropriate than only the ground state at zero temperature~\cite{sourlas1989spin, PhysRevLett.70.2968, belongie1994spin, nishimori2001statistical}. The condition for maximum restoration in error-correcting codes corresponds to the Nishimori condition known in the field of spin glasses~\cite{nishimori1981internal, le1988location, contucci2009spin}. In the case of maximum likelihood estimation, the ground truth $P_{\rm GT}$ influences the empirical distribution, but the two are different. Thus, it may be better to mix a probability model with a high value of the KL divergence to the ground truth than to select an optimal model. In this idea, the KL divergence plays the role of Hamiltonian in maximum likelihood estimation.

The structure of the present paper is as follows: in the next section, we introduce a gauge transformation that keeps the KL divergence to obtain the Nishimori condition for probability model estimation. In Section~\ref{sec_limit}, we will show that the probability model given by the Nishimori condition minimizes the KL divergence with the ground truth $P_{\rm GT}$ in the sense of expectation. The conventional maximum likelihood estimation selects an optimal probability model with the highest explanatory power for the given data from the prepared set of models. Our method requires a change in the concept of model selection from that in the conventional maximum likelihood method. We will discuss the two possible changes in Section~\ref{sec_interp}. The first is to select a model probabilistically, and the second is to construct a new model consisting of prepared probability models. Section~\ref{sec_example} presents two simple examples. The first is a model selection from a set formed by only two probability models. The second example is the set of probability models of the entire one-dimensional normal distributions. With prior knowledge that the ground truth $P_{\rm GT}$ is a normal distribution, it is intuitively expected that the best probability model obeys normal distribution. However, according to the latter interpretation described in Section~\ref{sec_interp}, we will show that the best probability model given by our method is not a normal distribution. Section~\ref{sec_imp} discusses improvements to the set of models. Maximum likelihood estimation requires a set of prepared probability models. Since the performance of maximum likelihood estimation depends on this set, improvement of the set is preferred. Section~\ref{sec_Beysian} discusses the relation between our proposed method and Beysian updating, which is standard for estimating probability distribution based on given data. We will discuss the sequential extension of the set. The last section will be devoted to a conclusion and future perspectives.

\section{Gauge invariance in KL divergence}

In our framework, the KL divergence between the empirical distribution and a probability model corresponds to a Hamiltonian in the theory of spin glasses. Under the Nishimori condition~\cite{nishimori1981internal, contucci2009spin} for spin glasses, the exact internal energy is calculated. To derive the Nishimori condition, which plays a central role in error-correcting codes and also in our discussion, let us consider the expected value of the KL divergence between the empirical distribution and a probability model corresponding to the internal energy in spin glasses: 
\begin{eqnarray}
\left\langle\left\langle D[P_{\rm emp}^\xi, P]\right\rangle_{\beta, \xi}\right\rangle_{\rm data}\,,
\end{eqnarray}
where $\left\langle\cdot\right\rangle_{\rm data}$ denotes the expected value with respect to the sample data $\xi=\{\xi_i\}_{i=1, \cdots, N}$, i.e., 
\begin{eqnarray}
\left\langle\cdot\right\rangle_{data} := \int\cdot\displaystyle\prod_{i=1}^N \left[d\xi_i P_{GT}(\xi_i)\right]\,.
\end{eqnarray}
The data $\xi$ is assumed to be sampled independently and identically from the ground truth $P_{\rm GT}$. In addition, $\left\langle\cdot\right\rangle_{\beta, \xi}$ represents Gibbs -- Boltzmann type weighted average with inverse temperature $\beta$. 
The Hamiltonian in our framework is the KL divergence between the empirical distribution $P_{\rm emp}^\xi (x) := \frac{1}{N}\sum_{i=1}^N \delta(x-\xi_i)$ and a probability model $P$: 
\begin{eqnarray}
\left\langle\cdot\right\rangle_{\beta, \xi} := \dfrac{\displaystyle\sum_{P\in\mathcal{M}}\cdot \exp\left\{-\beta D[P_{\rm emp}^\xi, P]\right\}}{\displaystyle\sum_{P\in\mathcal{M}}\exp\left\{-\beta D[P_{\rm emp}^\xi, P]\right\}}\,,
\end{eqnarray}
where $\sum_{P\in\mathcal{M}} \cdot$ is the sum over all models $P$ in the set $\mathcal{M}$ of probability models under consideration. 

In the argument of the Nishimori condition in spin glass theory, the gauge invariance plays an important role. Hamiltonian is invariant under the simultaneous transformation on the thermally fluctuating spins as stochastic variables and on the coupling constants as quenched random variables. In our case of the mixture of various probability models, the model $P$ corresponds to a stochastic variable. In contrast, since the data $\xi$ are generated randomly from $P_{\rm GT}$ and fixed once given, they can be considered quenched random. Let us introduce the following gauge transformation: 
\begin{eqnarray}
P(x) &\to& P_f(x) := \det\left(\nabla^T f(x)\right) P\left(f(x)\right)\,,\label{eq_transfP}\\
\xi_i &\to& \xi_i^f := f^{-1}(\xi_i)\,,
\end{eqnarray}
where $f$ is an arbitrary bijective map $\mathbb{R}^D \to \mathbb{R}^D$ and $f^{-1}$ is its inverse map. It is straightforward to show that $P_f(x)$ given by Eq.~(\ref{eq_transfP}) is indeed a density function. Since the empirical distribution is transformed under the gauge transformation as 
\begin{eqnarray}
P_{\rm emp}^{\xi^f}(x) &=& \dfrac{1}{N}\displaystyle\sum_{i=1}^N \delta\left(x-f^{-1}(\xi_i)\right)\nonumber\\
&=&\det\left(\nabla^T f(x)\right) P_{\rm emp}^\xi\left(f(x)\right)\,,\label{eq_emp}
\end{eqnarray}
we find 
\begin{eqnarray}
\dfrac{P_{\rm emp}^{\xi^f}(x)}{P_f(x)} = \dfrac{P_{\rm emp}^{\xi}\left(f(x)\right)}{P\left(f(x)\right)}\,.
\end{eqnarray}
In Eq.~(\ref{eq_emp}), we have used $\delta\left(f\left(x-f^{-1}(\xi_i)\right)\right) = 0$ when $f(x)\neq\xi_i$ since $f$ is a bijective map. In addition, $\nabla^T f(x)$ represents the Jacobian of $f(x)$. By the variable transformation $y=f(x)$, the KL divergence is shown to be invariant under the gauge transformation: 
\begin{eqnarray}
&&D[P_{\rm emp}^{\xi^f}, P_f]\nonumber\\
&=& \int dx\, \det\left(\nabla^T f(x)\right) P_{\rm emp}^\xi\left(f(x)\right)\ln\dfrac{P_{\rm emp}^\xi\left(f(x)\right)}{P\left(f(x)\right)}\nonumber\\
&=& \int dy\, P_{\rm emp}^\xi (y)\ln\dfrac{P_{\rm emp}^\xi (y)}{P(y)} = D[P_{\rm emp}^\xi, P]\,.
\end{eqnarray}
On the other hand, the probability of the ground truth for infinitesimal volume element $d\xi$ varies under the gauge transformation as 
\begin{eqnarray}
P_{\rm GT}(\xi) d\xi&\to& P_{\rm GT}(\xi^f) d\xi^f \nonumber\\
&=& \det\left(\nabla^T f^{-1}(\xi)\right) P_{\rm GT}\left(f^{-1}(\xi)\right)d\xi\nonumber\\
&:=& P_{\rm GT}^f(\xi)d\xi\,.\label{eq_Q}
\end{eqnarray}
Note that the following theorem holds~\cite{sei2011gradient, brenier1991polar, mccann1995existence}: for any pair $\left(p(x), q(x)\right)$ of density functions in $D$-dimension, there exists a convex function $\psi$ satisfying 
\begin{eqnarray}
p(x) = \left(\det\nabla\nabla^T \psi(x)\right) q\left(\nabla\psi(x)\right)\,,
\end{eqnarray}
where $\nabla\nabla^T \psi(x)$ denotes the Hessian of $\psi(x)$. The function $\psi$ is unique up to an arbitrary additive constant. Thus the target density $q(y)$ can be obtained from any density $p(x)$ by applying an appropriate variable transformation $y=\nabla\psi(x)$, and such a variable transformation is unique. In our framework, the gauge transformation with $f^{-1}(x)=\nabla\psi(x)$ uniquely determines $P_{\rm GT}^f$ connected to the ground truth $P_{\rm GT}$. Using the invariance of the KL divergence under the gauge transformation, we find 
\begin{eqnarray}
&&\left\langle\left\langle D[P_{\rm emp}^\xi, P]\right\rangle_{\beta, \xi}\right\rangle_{\rm data}
= \int\displaystyle\prod_{i=1}^N\left[d\xi_i\, P_{\rm GT}(\xi_i)\right]\dfrac{D_{\mathcal{M}, I}^{\xi, \beta}}{Z_{\mathcal{M}, I}^{\xi, \beta}}\nonumber\\
&&=\int\displaystyle\prod_{i=1}^N\left[d\xi^f_i P_{\rm GT}(\xi^f_i)\right]\dfrac{D_{\mathcal{M}, f}^{\xi, \beta}}{Z_{\mathcal{M}, f}^{\xi, \beta}}\nonumber\\
&&=\int\displaystyle\prod_{i=1}^N\left[d\xi^f_i P_{\rm GT}(\xi^f_i)\right]\dfrac{D_{\mathcal{M}, I}^{\xi, \beta}}{Z_{\mathcal{M}, I}^{\xi, \beta}}\,,
\label{eq_Dgauge}
\end{eqnarray}
where $D^{\xi, \beta}_{\mathcal{M}, f} := \sum_{P_f\in\mathcal{M}_f}D[P_{\rm emp}^{\xi^f}, P_f] e^{-\beta D[P_{\rm emp}^{\xi^f}, P_f]}$ and $Z_{\mathcal{M}, f}^{\xi, \beta} := \sum_{P_f\in\mathcal{M}_f}e^{-\beta D[P_{\rm emp}^{\xi^f}, P_f]}$ with the notation $\mathcal{M}_f$ denoting the image of $\mathcal{M}$ by $f$. The relations $P_{\rm emp}^{\xi^I} = P_{\rm emp}^\xi$, $P_I = P$, and $\mathcal{M}_I = \mathcal{M}$ have been used for an identity map $I$. In the last equality in Eq.~(\ref{eq_Dgauge}), the expression has been rewritten as a sum over $P\in\mathcal{M}$ since each $P_f\in\mathcal{M}_f$ corresponds to an appropriate $P\in\mathcal{M}$ by the gauge transformation. In addition, since an arbitrary density $P_{\rm GT}^f(\xi)$ can be represented as 
\begin{eqnarray}
P_{\rm GT}^f (\xi) = \exp\left[\int dx\, \delta(x-\xi)\ln P_{\rm GT}^f(x)\right]\label{eq_Pxi}\,,
\end{eqnarray}
and the formal expression 
\begin{eqnarray}
\displaystyle\prod_{i=1}^N e^{\int dy\, \delta(y-\xi_i)\ln P_{\rm GT}^f(y)} = e^{N\int dy\, P_{\rm emp}^\xi (y)\ln P_{\rm GT}^f(y)}\label{eq_deltaEmp}
\end{eqnarray}
holds, Eq.~(\ref{eq_Dgauge}) can be rewritten through Eqs.~(\ref{eq_Q}), (\ref{eq_Pxi}), and (\ref{eq_deltaEmp}) as 
\begin{eqnarray}
\left\langle\left\langle D[P_{\rm emp}^\xi, P]\right\rangle_{\beta, \xi}\right\rangle_{\rm data} = 
\int\displaystyle\prod_{i=1}^N d\xi_i\, E^{\xi, N}_{P_{\rm GT}^f}
\dfrac{D_{\mathcal{M}, I}^{\xi, \beta}}{Z_{\mathcal{M}, I}^{\xi, \beta}}\,,
\end{eqnarray}
where $E^{\xi, \beta}_P := e^{\beta \int dy\, P_{\rm emp}^{\xi}(y)\ln P(y)}$. Since $\int dx\, P_{\rm emp}^\xi (x)\ln P_{\rm emp}^\xi (x)$ in the definition of the KL divergence is independent of the model $P$, we have 
\begin{eqnarray}
&&\left\langle\left\langle D[P_{\rm emp}^\xi, P]\right\rangle_{\beta, \xi}\right\rangle_{\rm data} \nonumber\\
&&= \int\displaystyle\prod_{i=1}^N d\xi_i\, E_{P_{\rm GT}^f}^{\xi, N}
\dfrac{\displaystyle\sum_{P\in\mathcal{M}} D[P_{\rm emp}^\xi, P]E_P^{\xi, \beta}}{\displaystyle\sum_{P\in\mathcal{M}} E_P^{\xi, \beta}}\,.\label{eq_Destimate}
\end{eqnarray}

Since $\left\langle\left\langle D[P_{\rm emp}^\xi, P]\right\rangle_{\beta, \xi}\right\rangle_{\rm data}$ is invariant under the gauge transformation, the result of Eq.~(\ref{eq_Destimate}) is independent of $f$. Thus the result is the same as the average over all elements $f$ of an arbitrary set of bijective maps $\mathcal{F}$. Denoting the number (or volume in the case of a continuum) of the bijective maps by $N_{\mathcal{F}}$, we obtain 
\begin{widetext}
\begin{eqnarray}
\left\langle\left\langle D[P_{\rm emp}^\xi, P]\right\rangle_{\beta, \xi}\right\rangle_{\rm data}
= \dfrac{1}{N_{\mathcal{F}}}\int\displaystyle\prod_{i=1}^N d\xi_i\,
\left[\displaystyle\sum_{f\in\mathcal{F}} E_{P_{\rm GT}^f}^{\xi, N}\right]
\dfrac{\displaystyle\sum_{P\in\mathcal{M}} D[P_{\rm emp}^\xi, P]E_P^{\xi, \beta}}{\displaystyle\sum_{P\in\mathcal{M}} E_P^{\xi, \beta}}\,.
\end{eqnarray}
\end{widetext}
Using the correspondence between $f$ and $P_{\rm GT}^f$, the sum for $f$ becomes the sum for $P_{\rm GT}^f$. Then we have 
\begin{widetext}
\begin{eqnarray}
\left\langle\left\langle D[P_{\rm emp}^\xi, P]\right\rangle_{\beta, \xi}\right\rangle_{\rm data}
= \dfrac{1}{N_{\mathcal{M_{\rm GT}^\mathcal{F}}}}\int\displaystyle\prod_{i=1}^N d\xi_i
\left[\displaystyle\sum_{P\in\mathcal{M}} D[P_{\rm emp}^\xi, P]E_P^{\xi, \beta}\right]
\dfrac{\displaystyle\sum_{P\in\mathcal{M}_{\rm GT}^\mathcal{F}} E_P^{\xi, N}}{\displaystyle\sum_{P\in\mathcal{M}}E_P^{\xi, \beta}}\,,\label{eq_ratio}
\end{eqnarray}
\end{widetext}
where $\mathcal{M}_{\rm GT}^\mathcal{F}$ is the set composed of all of $P_{\rm GT}^f$ given by the bijective maps $f\in\mathcal{F}$, and $N_{\mathcal{M}_{\rm GT}^\mathcal{F}}$ is the number of its elements (or volume in the case of a continuum). Note that $N_\mathcal{F}=N_{\mathcal{M}_{\rm GT}^\mathcal{F}}$ holds when $\mathcal{F}$ is a discrete set. The ratio of Boltzmann weights appearing on the right-hand side in Eq.~(\ref{eq_ratio}) becomes unity when $\beta=N$ and $\mathcal{M}_{\rm GT}^\mathcal{F} = \mathcal{M}$. The condition for the unity in the ratio is called the Nishimori condition hereafter. 

The condition for $\mathcal{M}_{\rm GT}^\mathcal{F}=\mathcal{M}$ is given as follows. We can use an arbitrary bijective map $f$ in the gauge transformation. Since $P_{\rm GT}^f (x) = \det\left(\nabla^T f^{-1}(x)\right)P_{\rm GT}\left( f^{-1}(x)\right)$ and, for any density function $Q$, there exists a unique bijective map $g$ satisfying $Q(x) = \det\left( \nabla^T g(x)\right) P_{\rm GT} \left( g(x) \right)$, we can find a unique bijective map $f$ satisfying $P_{\rm GT}^f (x) = P(x)$ for each model $P\in\mathcal{M}$. Thus, assuming that $\mathcal{F}$ is the set consisting of all bijective maps $f$ satisfying $P_{\rm GT}^f (x) = P(x)\in\mathcal{M}$, $\mathcal{M}_{\rm GT}^\mathcal{F}=\mathcal{M}$ with $N_\mathcal{F}=N_\mathcal{M}$ holds, where $N_\mathcal{M}$ is the number of elements of $\mathcal{M}$ (or volume in the case of a continuum). Since $P_{\rm GT}$ is unknown, to give a concrete set of the bijective maps $\mathcal{F}$ is impossible. However, such a set exists, and a concrete expression of $\mathcal{F}$ is not required. Therefore, for an arbitrary set of probability models $\mathcal{M}$, there exists an appropriate set of bijective maps $\mathcal{F}$ that allows the ratio of Boltzmann weights appearing on the right-hand side of Eq.~(\ref{eq_ratio}) to be set unity. The Nishimori condition is always satisfied by $\beta=N$ and $\mathcal{M}_{\rm GT}^\mathcal{F} = \mathcal{M}$. In this case, Eq.~(\ref{eq_ratio}) yields the simple form as 
\begin{eqnarray}
&&\left\langle\left\langle D[P_{\rm emp}^\xi, P]\right\rangle_{N, \xi}\right\rangle_{\rm data}\nonumber\\
&=&\dfrac{1}{N_\mathcal{M}}\displaystyle\sum_{P\in\mathcal{M}}\int\prod_{i=1}^N\left[d\xi_i\, P(\xi_i)\right] D[P_{\rm emp}^\xi, P]\,.\label{eq_internalEnergy}
\end{eqnarray}
From this expression, $\left\langle\left\langle D[P_{\rm emp}^\xi, P]\right\rangle_{N, \xi}\right\rangle_{\rm data} $ is independent of $P_{\rm GT}$, and is given as the average over $P\in\mathcal{M}$ of the expected KL divergence between the model $P$ and the empirical distribution given by the realizations of the data $\{\xi_i\}_{i=1, \cdots, N}$ generated from $P$. In other words, the expectation for the KL divergence is calculated only from the probability models in principle on the Nishimori condition. It is consistent with the fact that the internal energy is calculated exactly on the Nishimori condition in the framework of spin glass theory~\cite{nishimori2001statistical}. 

In the framework of AIC, the unbiased estimator for the KL divergence between the empirical distribution $P_{\rm emp}^\xi$ and the probability model $P$ is evaluated specifically~\cite{akaike1974new, konishi2008information}. Correspondingly, it has been shown that $\left\langle\left\langle D[P_{\rm emp}^\xi, P]\right\rangle_{N, \xi}\right\rangle_{\rm data}$ can be calculated in principle in our case. 
However, as seen in the next section, such a concrete expression is not required for the best estimation of $P_{\rm GT}$. Note that $\left\langle\left\langle D[P_{\rm emp}^\xi, P]\right\rangle_{N, \xi}\right\rangle_{\rm data}$ depends only on the set $\mathcal{M}$ of prepared probability models. Since $P_{\rm GT}$ is unknown, and we estimate using only the sample $\{\xi_i\}_{i=1, \cdots, N}$, it is natural that the ground truth $P_{\rm GT}$ does not explicitly affect $\left\langle\left\langle D[P_{\rm emp}^\xi, P]\right\rangle_{N, \xi}\right\rangle_{\rm data}$, which is irrelevant to estimation accuracy. 

\section{Performance inequality}\label{sec_limit}

In the present section, we will confirm that $\beta=N$ gives the best estimation in our framework, in which an average model weighted by KL divergence provides the estimation. In other words, we will show that the expectation of the KL divergence between the grand truth $P_{\rm GT}$ and the estimated density $\hat{P}_\beta^\xi$, i.e., $\left\langle D[P_{\rm GT}, \hat{P}_{\beta}^\xi]\right\rangle_{\rm data}$ becomes minimum at $\beta = N$, where the estimation $\hat{P}_\beta^\xi$ is an averaged model weighted by the KL divergence to the empirical distribution given by the data $\{\xi_i\}_{i=1, \cdots, N}$ with the inverse temperature $\beta$: 
\begin{eqnarray}
\hat{P}_\beta^\xi (x) = \dfrac{\displaystyle\sum_{P\in\mathcal{M}} P(x)e^{-\beta D[P_{\rm emp}^\xi, P]}}{\displaystyle\sum_{P\in\mathcal{M}} e^{-\beta D[P_{\rm emp}^\xi, P]}}\,.
\end{eqnarray}
Considering the difference between $\left\langle D[P_{\rm GT}, \hat{P}_{\beta}^\xi]\right\rangle_{\rm data}$ and $\left\langle D[P_{\rm GT}, \hat{P}_{N}^\xi]\right\rangle_{\rm data}$, we have 
\begin{eqnarray}
&&\left\langle D[P_{\rm GT}, \hat{P}_{\beta}^\xi]\right\rangle_{\rm data} - \left\langle D[P_{\rm GT}, \hat{P}_{N}^\xi]\right\rangle_{\rm data} \nonumber\\
&=&-\int\displaystyle\prod_{i=1}^N \left[d\xi_i P_{\rm GT}(\xi_i)\right]\int dx\, P_{\rm GT}(x)\ln\dfrac{\hat{P}_\beta^\xi (x)}{\hat{P}_N^\xi (x)}\,.
\label{eq_diff}
\end{eqnarray}
Note that the ground truth changes as 
\begin{eqnarray}
&&\displaystyle\prod_{i=1}^N\left[d\xi_i P_{\rm GT}(\xi_i)\right]P_{\rm GT}(x) 
\to \prod_{i=1}^N d\xi_i\,P_{\rm GT}^f(x) E_{P_{\rm GT}^f}^{\xi, N}\nonumber\\
&&
\end{eqnarray}
under the gauge transformation, while $\hat{P}_\beta^\xi / \hat{P}^\xi_N$ is invariant. Since the result in Eq.~(\ref{eq_diff}) does not change before and after the gauge transformation, it equals the average over all gauges $f\in\mathcal{F}$. Since $\mathcal{F}$ is the set of bijective maps connecting $P_{\rm GT}$ and $P\in\mathcal{M}$ as $P_{\rm GT}^f (x)=P(x)$ in our framework, the average over $f\in\mathcal{F}$ equals the average over $P\in\mathcal{M}$. Then we find 
\begin{eqnarray}
&&-\int\displaystyle\prod_{i=1}^N \left[d\xi_i P_{\rm GT}(\xi_i)\right]\int dx\, P_{\rm GT}(x)\ln\dfrac{\hat{P}_\beta^\xi (x)}{\hat{P}_N^\xi (x)}\nonumber\\
&=&-\dfrac{1}{N_\mathcal{M}}\int\displaystyle\prod_{i=1}^N d\xi_i\int dx\, S_{\mathcal{M}}^{\xi, N}(x)\ln\dfrac{\hat{P}_\beta^\xi (x)}{\hat{P}_N^\xi (x)}\nonumber\\
&\ge& -\dfrac{1}{N_\mathcal{M}}\int\displaystyle\prod_{i=1}^N d\xi_i\int dx\, S_{\mathcal{M}}^{\xi, N}(x)\left[\dfrac{\hat{P}_\beta^\xi (x)}{\hat{P}_N^\xi (x)}-1\right]\,,\label{eq_ineq}
\end{eqnarray}
where $S_{\mathcal{M}}^{\xi, \beta} (x):= \sum_{P\in\mathcal{M}}P(x)E^{\xi, \beta}_P$, and $\ln z \ge z-1$ for $z > 0$ has been used. 

Using the definition of the KL divergence $D[P^\xi_{\rm emp}, P]$, the concrete form of $\hat{P}_\beta^\xi(x)$ is rewritten as  
\begin{eqnarray}
\hat{P}^\xi_\beta(x)=\dfrac{S_{\mathcal M}^{\xi, \beta}(x)}{\displaystyle\sum_{P\in{\mathcal M}} E_P^{\xi, \beta}}\,.
\end{eqnarray}
Thus we evaluate the last line in Eq.~(\ref{eq_ineq}) as 
\begin{eqnarray}
-\dfrac{1}{N_\mathcal{M}}\int\displaystyle\prod_{i=1}^N d\xi_i\int dx\, 
\left[S_{\mathcal{M}}^{\xi, \beta}(x)
\dfrac{\displaystyle\sum_{P\in\mathcal{M}} E_P^{\xi, N}}{\displaystyle\sum_{P\in\mathcal{M}}E_P^{\xi, \beta}} - S_{\mathcal{M}}^{\xi, N}(x)\right]\,.\nonumber\\
&&\label{eq_performance}
\end{eqnarray}
Since $\int dx\, P(x)=1$ holds for an arbitrary model $P$, we have 
\begin{eqnarray}
\int dx\, S_\mathcal{M}^{\xi, \beta}(x) = \displaystyle\sum_{P\in\mathcal{M}} E_P^{\xi, \beta}
\end{eqnarray}
for arbitrary $\beta$. Using this fact, we find that Eq.~(\ref{eq_performance}) vanishes. Therefore we attain 
\begin{eqnarray}
\left\langle D[P_{\rm GT}, \hat{P}_\beta^\xi]\right\rangle_{\rm data} \ge \left\langle D[P_{\rm GT}, \hat{P}_N^\xi]\right\rangle_{\rm data}\,.
\end{eqnarray}
The density function $\hat{P}_N^\xi$ estimated on the Nishimori condition is closest to the ground truth $P_{GT}$ in the sense of expectation.

The above result is suggestive. The most accurate estimation in the limit of $N\to\infty$ corresponds to the Nishimori temperature $\beta\to\infty$, or zero temperature estimation. It is consistent with the well-known fact that the best model is obtained through the conventional maximum likelihood method when the sample size is sufficiently large. In such a case, the criterion for model selection is minimization for the KL divergence to the empirical distribution $P_{\rm emp}^\xi$ determined from data $\{\xi_i\}_{i=1, \cdots, N}$. In the conventional maximum likelihood method, a new model $P$ with lower KL divergence replaces an old one with higher KL divergence. On the other hand, in our approach, we weigh these old models by the KL divergence to obtain a more accurate model. 

To reduce the value of $\left\langle D[P_{\rm GT}, \hat{P}_N^\xi]\right\rangle_{\rm data}$, we should increase the sample size $N$ or improve the set of models $\mathcal{M}$ under consideration. Note that $\hat{P}_N^\xi$ does not depend on $P_{\rm GT}$ but on the set of models $\mathcal{M}$. In other words, the same $\mathcal{M}$ yields the same estimation result $\hat{P}_N^\xi$, even if the same sample $\{\xi_i\}_{i=1,\cdots, N}$ is generated from different ground truths. It is a natural result because our estimation bases only on the data. The choice of $\mathcal{M}$ determines the estimation accuracy. In Section~\ref{sec_imp}, we will discuss the improvement of $\mathcal{M}$. $\left\langle D[P_{\rm GT}, \hat{P}_N^\xi]\right\rangle_{\rm data}$ is an indicator for the estimation accuracy, and depends on $P_{\rm GT}$, although $\hat{P}_N^\xi$ is independent of $P_{\rm GT}$. We have shown that $\left\langle D[P_{\rm GT}, \hat{P}_\beta^\xi]\right\rangle_{\rm data}$ gives the minimum at $\beta = N$ using gauge invariance for the difference in Eq.~(\ref{eq_diff}). However, since $D[P_{\rm GT}, \hat{P}_\beta^\xi]$ is not gauge invariant, the value of $\left\langle D[P_{\rm GT}, \hat{P}_\beta^\xi]\right\rangle_{\rm data}$ depends on $P_{\rm GT}$. 

\section{Two possible interpretation}\label{sec_interp}

We will show in Section~\ref{sec_example} that the estimation $\hat{P}_N^\xi$ by our method is generally not in the prepared set $\mathcal{M}$, i.e., $\hat{P}_N^\xi\notin\mathcal{M}$. In this case, the best model does not belong to the initially prepared set of models. On the other hand, the conventional maximum likelihood method selects the model most explanatory to the realized data from the initially prepared models. Thus our framework requires a change in the interpretation of maximum likelihood estimation. There are two possible interpretations of our method. The first is to select the model probabilistically. In our approach, the prepared models have Boltzmann weights. We can select a model probabilistically according to the Boltzmann weight if only one of the models in $\mathcal{M}$ is allowed as an estimation result. In this case, a lower KL divergence is achieved compared to one obtained by the conventional maximum likelihood method in the sense of expectation. The second is to accept $\hat{P}_N^\xi\notin\mathcal{M}$ as a probability model. In other words, we construct a new probability model from the set of initially prepared models $P\in\mathcal{M}$. In this case, the best estimation is deterministic. If we have prior knowledge on $P_{\rm GT}$, the initial set of models $\mathcal{M}$ is naturally prepared based on prior knowledge. However, the newly constructed model $\hat{P}_N^\xi$ does not generally belong to the set of models consistent with the prior knowledge. We will discuss such an example in Section~\ref{sec_example}. 

\section{Simple examples}\label{sec_example}

Based on the discussion up to Section~\ref{sec_limit}, the best estimation for the given data $\{\xi_i\}_{i=1, \cdots, N}$ in our framework $\hat{P}_N^\xi (x)$ is given as 
\begin{eqnarray}
\hat{P}_N^\xi(x) = \dfrac{\displaystyle\sum_P P(x)\prod_{i=1}^N P(\xi_i)}{\displaystyle\sum_P \prod_{i=1}^N P(\xi_i)}\,.\label{eq_Phat}
\end{eqnarray}
In the present section, we deal with simple examples in which $\hat{P}_N^\xi$ is tractable and discuss the implications of our method.

\subsection{Model selection from two}

Let us estimate the probability density behind the given one-dimensional data $\{\xi_i\}_{i=1, \cdots, N}$. We initially prepare two one-dimensional normal distributions $P_{\pm}(x) := \exp\left[-(x\mp a)^2/2\right]/\sqrt{2\pi}$ ($a>0$). In other words, we set $\mathcal{M} = \{P_+, P_-\}$. 
In this case, simple calculation yield 
\begin{eqnarray}
\hat{P}_N^\xi (x) &=& \dfrac{e^{-\frac{1}{2}(x^2+a^2)}}{\sqrt{2\pi}}\left[\cosh ax + \left(\sinh ax\right)\left(\tanh aN\bar{\xi}\right)\right]\,,\nonumber\\
&&
\end{eqnarray}
where $\bar{\xi} := \frac{1}{N}\sum_{i=1}^N \xi_i$ is the sample average. 
Since $\tanh aN\bar{\xi}\to{\rm sgn}\,\bar{\xi}$ in the limit of $N\to\infty$ with $a>0$, we find $\hat{P}_N^\xi (x)\to \exp\left[-\frac{1}{2}\left(x-a{\rm sgn}\,\bar{\xi}\right)^2\right]/\sqrt{2\pi}$ in large $N$ limit. Thus our method selects one from the two models depending on $\bar{\xi}$ without uncertainty for $N\to \infty$.

\begin{figure}[b]
\centering
\includegraphics[keepaspectratio, scale=0.5]{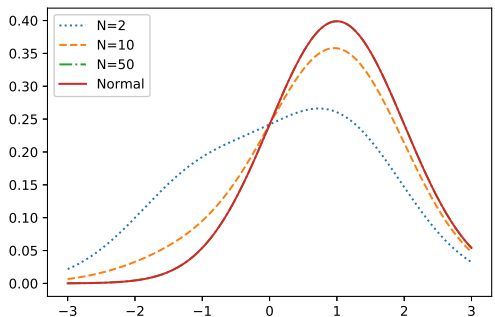}
\caption{\label{fig_selection} (Color online.) $\hat{P}_N^\xi (x)$ obtained for a set of two normal distributions. The results for $N=2$, $N=10$, and $N=50$ are represented by dotted, dashed, and single-point dashed curves, respectively. The solid curve depicts the normal distribution $P_+ (x) = \exp\left[-\frac{1}{2}(x-a)^2\right]/\sqrt{2\pi}$. The curves for $\hat{P}_N^\xi$ with $N=50$ and $P_+$ are almost overlapped. The parameters are set to be $\bar{\xi}=0.1$ and $a=1$.}
\end{figure}

As shown in Fig.~\ref{fig_selection}, $\hat{P}_N^\xi (x)$ differs from the normal distribution for small $N$ and converges rapidly to the normal distribution as $N$ increases. The deviation from the normal distribution for small $N$ can be interpreted as an interference between the two models. Since we cannot select one model definitely for a sample with a finite size, additional variance emerges due to the other model.  

In addition to the effect of finite $N$, it is interesting that $\hat{P}_N^\xi$ is given even for $\bar{\xi}=0$. In the conventional maximum likelihood method, the KL divergence from the empirical distribution to $P_+$ equals the KL divergence to $P_-$ in such a situation. 
Then one appropriate model cannot be selected as the solution to the optimization problem. On the other hand, a unique optimum solution becomes $\hat{P}_N^\xi (x) = \frac{1}{2}\left[P_+(x) + P_-(x)\right]$ in our framework. Although the estimated model does not belong to the initially prepared set of probability models, i.e., $\hat{P}_N^\xi\notin\mathcal{M} = \{P_+, P_-\}$, it is considered as an intuitively appropriate solution to the optimization problem.

\subsection{Estimation from data generated by a normal distribution}

In the present subsection, we assume the case where the ground truth $P_{\rm GT}$ is known to be a one-dimensional normal distribution. In such a situation, the natural set of models $\mathcal{M}$ consists of all possible one-dimensional normal distributions, i.e., $P(x|\sigma, a) = \frac{1}{\sqrt{2\pi\sigma^2}}\exp\left[-\frac{1}{2\sigma^2}(x-a)^2\right]$ defined for all positive numbers $\sigma >0$ and real numbers $a\in\mathbb{R}$. 

\begin{figure}[b]
\centering
\includegraphics[keepaspectratio, scale=0.5]{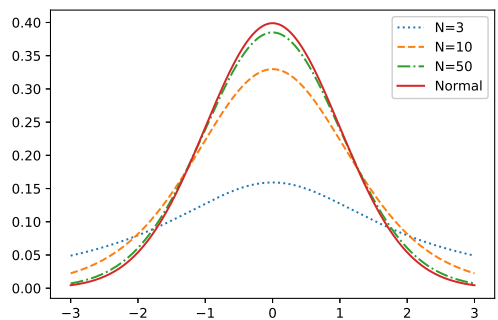}
\caption{\label{fig_deformedNormal} (Color online.) $\hat{P}_N^\xi (x)$ for $\mathcal{M}$ consisting of all normal distributions. The results for $N=3$, $N=10$, and $N=50$ are represented by dotted, dashed, and single-point dashed curves, respectively. The solid curve depicts the normal distribution $\exp\left(-\frac{1}{2}x^2\right)/\sqrt{2\pi}$. The parameters are set to be $\bar{\xi}=0$ and $V_\xi =1$.}
\end{figure}

Defining the set of probability models by $\mathcal{M} := \{P(x|\sigma, a) | \sigma > 0, a\in\mathbb{R}\}$, we evaluate the denominator in Eq.~(\ref{eq_Phat}) as 
\begin{eqnarray}
Z_{\mathcal{M}}^\xi
=\dfrac{1}{2}C_N U_N\left(V_\xi\right)\Gamma\left(\dfrac{N}{2}-1\right)\,,
\end{eqnarray}
where $V_\xi := \frac{1}{N}\sum_{i=1}^N \xi_i^2 - \bar{\xi}^2$ is the sample variance for the data $\xi$, and $\Gamma(z) := \int_0^\infty dt\, t^{z-1}e^{-t}$ is a gamma function. We have used the notation 
 $Z_{\mathcal{M}}^\xi := \sum_{P\in\mathcal{M}}\prod_{i=1}^N P(\xi_i)$. 
In addition, the notations $C_N := \left[(2\pi)^{(N-1)/2}\sqrt{N}\right]^{-1}$ and $U_N(z) := \left[2/(Nz)\right]^{N/2 - 1}$ have been used. 
Similarly, the numerator in Eq.~(\ref{eq_Phat}) is 
\begin{eqnarray}
S_{\mathcal{M}}^{\xi}
= \dfrac{1}{2}
C_{N+1}U_{N+1}\left(V_{\xi, x}\right)\Gamma\left(\dfrac{N}{2}-\dfrac{1}{2}\right)\,,
\end{eqnarray}
where 
$V_{\xi, x} := \frac{1}{N+1}\left(N\bar{\xi} + x\right)^2 - \left[\frac{1}{N+1}\left(N\bar{\xi} + x\right)\right]^2$
is the sample variance for the data composed of $\xi$ and the new sample point $x$. The notation $S_\mathcal{M}^\xi (x) := S_\mathcal{M}^{\xi, N}$ has also been used. Then the best estimation on the Nishimori condition becomes 
\begin{eqnarray}
\hat{P}_N^\xi (x) =
\dfrac{C_{N+1} U_{N+1}\left(V_{\xi, x}\right)}{C_N U_{N}\left( V_\xi\right)}
\dfrac{\Gamma(N/2-1/2)}{\Gamma(N/2 - 1)}
\,.\label{eq_finN}
\end{eqnarray}
Since 
\begin{eqnarray}
\hat{P}_N^\xi (x)\propto\left[\dfrac{(N+1) V_\xi}{\left( x-\bar{\xi}\right)^2+(N+1)V_\xi}\right]^{(N-1)/2}\,,
\end{eqnarray}
it is found that the $\hat{P}_N^\xi$ is given by $(N-1)/2$ power of the Lorentzian distribution with the peak located at $\bar{\xi}$ and the peak width of $\left[(N+1) V_\xi\right]^{1/2}$ as shown in Fig.~\ref{fig_deformedNormal}. The obtained model is not a normal distribution, although our estimation assumes that the ground truth $P_{\rm GT}$ obeys normal distribution. This result seems to be counterintuitive. However, the prior knowledge of the ground truth of normal distribution does not limit the model to normal distribution. When the sample size $N$ is finite, selecting one normal distribution may lead to a large deviation from the ground truth $P_{\rm GT}$. To be ruled out the possibility of a large deviation, the best estimation deviates from any normal distributions. We emphasize that such a situation occurs because of the finite sample size $N$. Indeed, we find in the limit $N\to\infty$ that 
\begin{eqnarray}
\hat{P}_N^\xi (x)\to\dfrac{1}{\sqrt{2\pi V_\xi}}\exp\left[-\dfrac{1}{2V_\xi}\left( x-\bar{\xi}\right)^2\right]
\end{eqnarray}
holds using Stirling's formula $\Gamma(z+1)\sim\sqrt{2\pi z}\left(z/e\right)^z$ and the definition of exponential function $(1+z/n)^n \to e^z$ for $n\to\infty$. Thus, in the limit of large samples, our result reproduces the normal distribution, which the prior knowledge implicates, and is consistent with the well-known conclusion of the maximum likelihood estimation. $\hat{P}_N^\xi$ given by Eq.~(\ref{eq_finN}) can be interpreted as a finite size effect version of normal distribution. 

\section{Improvement of model set}\label{sec_imp}

The best estimation $\hat{P}_N^\xi$ depends only on the set of prepared models $\mathcal{M}$. Let us consider the sequential extension of the set of models $\mathcal{M}$ to improve the estimation accuracy. In this process, a new set of models $\mathcal{M}_\ast$ is constructed by the addition of a new model $P_\ast$ to the original set $\mathcal{M}$. If the number $N_\mathcal{M}$ of elements in $\mathcal{M}$ is sufficiently large, the constructed model $\hat{P}_{N, \ast}^\xi$ from the modified set $\mathcal{M}_\ast$ is expressed as 
\begin{widetext}
\begin{eqnarray}
\hat{P}_{N, \ast}^\xi (x) =
\hat{P}_N^\xi (x) + \left[P_\ast (x) - \hat{P}_N^\xi (x)\right]\dfrac{z_\ast^\xi}{Z_\mathcal{M}^\xi}
+ \mathcal{O}\left(\left( z_\ast^\xi / Z_\mathcal{M*}^\xi\right)^2\right)\,,
\end{eqnarray}
\end{widetext}
where $z_\ast^\xi := \prod_{i=1}^N P_\ast (\xi_i)$. For practical purposes, preferable $\hat{P}_{N, \ast}^\xi$ is closer to the ground truth $P_{\rm GT}$ than $\hat{P}_N^\xi$ in the KL divergence. The expectation of the KL divergence between the newly obtained model $\hat{P}_{N, \ast}^\xi$ and $P_{\rm GT}$ is calculated as 
\begin{widetext}
\begin{equation}
\left\langle D[P_{\rm GT}, \hat{P}_{N, \ast}^\xi]\right\rangle_{\rm data}
= \left\langle D[P_{\rm GT}, \hat{P}_N^\xi]\right\rangle_{\rm data}
- \int\displaystyle\prod_{i=1}^N\left[ d\xi_i\, P_{\rm GT}(\xi_i)\right]\int dx
\left[P_\ast (x) - \hat{P}_N^\xi (x)\right]
\frac{P_{\rm GT}(x) z_\ast^\xi}{\hat{P}_N^\xi (x) Z_\mathcal{M}^\xi}
+ \mathcal{O}\left(\left(z_\ast^\xi / Z_\mathcal{M}^\xi \right)^2\right)\,.
\end{equation}
\end{widetext}
Since $\hat{P}_N^\xi (x)$ is non-negative, it is necessary to add $P_\ast$ such that $P_\ast (x) - \hat{P}_N^\xi (x)$ becomes as large as possible at $x$ where $P_{\rm GT}(x)/\hat{P}_N^\xi (x)$ is large to make $\left\langle D[P_{\rm GT}, \hat{P}_{N, \ast}^\xi]\right\rangle_{\rm data}$ as small as possible. In other words, it is preferable to add a model that assigns as high probability as possible at $x$, which is underestimated in the originally constructed model $\hat{P}_N^\xi$. The systematic expansion of $\mathcal{M}$ described above leads to a decrease in the expected value of the KL divergence between $\hat{P}_N^\xi$ and $P_{\rm GT}$. Such a systematic process may yield an algorithm that can efficiently estimate $P_{\rm GT}$.

\section{Relation to Bayesian updating}\label{sec_Beysian}

Our proposed method is related to Bayesian updating. It updates the prior distribution $\hat{p}(\theta)$ for the parameter $\theta$ of the model $P_\theta(x)$ based on observed data: 
\begin{eqnarray}
\hat{p}(\theta|\xi) = \dfrac{P_\theta(\xi)\hat{p}(\theta)}{\displaystyle\sum_\theta P_\theta(\xi)\hat{p}(\theta)}\,.
\end{eqnarray}
In our method, the weights for probability models are updated when a new data $\xi_{N+1}$ is observed. According to our result Eq.~(\ref{eq_Phat}), the weight for the model $P$ is given as 
\begin{eqnarray}
\hat{p}(P|\xi)=\dfrac{\displaystyle\prod_{i=1}^N P(\xi_i)}{\displaystyle\sum_P \prod_{i=1}^N P(\xi_i)}\label{eq_fixed}
\end{eqnarray}
when the realization of the data $\{\xi_i\}_{i=1, \cdots, N}$ is observed. 
Interpreting $\hat{P}(P|\xi_1, \cdots, \xi_N)$ as the posterior distribution of model the $P$, when new data $\xi_{N+1}$ are added, the posterior distribution of the model $P$ is given by Bayesian updating as 
\begin{eqnarray}
&&\hat{p}(P|\xi_1, \cdots, \xi_N, \xi_{N+1}) \nonumber\\
&=& \dfrac{P(\xi_{N+1})\hat{p}(P|\xi_1, \cdots, \xi_N)}{\displaystyle\sum_P P(\xi_{N+1})\hat{p}(P |\xi_1, \cdots, \xi_N)}\,.\label{eq_update}
\end{eqnarray}
Note that the update rule Eq.~(\ref{eq_update}) is a direct result from Eq.~(\ref{eq_Phat}). Our proposed method is concluded to be consistent with Bayesian updating starting from a prior distribution of equal weights for all probability models.

By repeating the Bayesian updating with given data $\{\xi\}_{i=1, \cdots, N}$ fixed, we can improve the posterior distribution for the probability model $P$ by obtaining the posterior distribution. In this approach, the posterior distribution for the model $P$ of the $(k+1)^{th}$ step $\hat{p}_{k+1}(P|\xi_1, \cdots, \xi_N)$ is obtained from the posterior distribution of the $k^{th}$ step $\hat{p}_k(P|\xi_1, \cdots, \xi_N)$ as the prior distribution. $\hat{p}_{k+1}(P|\xi_1, \cdots, \xi_N)$ is updated as 
\begin{eqnarray}
\hat{p}_{k+1}(P|\xi_1, \cdots, \xi_N) = \dfrac{\hat{p}_k(P|\xi_1, \cdots, \xi_N)\displaystyle\prod_{i=1}^N P(\xi_i)}{\displaystyle\sum_P \hat{p}_k(P|\xi_1, \cdots, \xi_N)\prod_{i=1}^N P(\xi_i)}\,,\nonumber\\
\end{eqnarray}
where $\hat{p}_0(P|\xi_1, \cdots, \xi_N)=\hat{p}_0(P)$ is the initial prior distribution for the model $P$. For this update rule, the fixed point $\hat{p}_\infty(P|\xi_1, \cdots, \xi_N)$ is given by Eq.~(\ref{eq_fixed}). Thus our method can be understood as a fixed point of Bayesian updating.

It has been pointed out that Bayesian updating does not require regularization. We have found that Bayesian updating can be interpreted as an error-correcting version of the maximum likelihood method. In other words, we can conclude that Bayesian updating is an estimation method that minimizes the KL divergence between the ground truth distribution and the probability model in the sense of expected value. 

\section{Conclusion}

We have discussed a method for estimating density functions based on gauge symmetry in the KL divergence. Our framework naturally prevents overfitting, although the conventional maximum likelihood method can lead to incorrect estimation results due to overfitting to empirical distributions. While empirical regularization to prevent overfitting often requires a search for the optimal hyperparameters, our method does not require adjustment of hyperparameters because the optimal hyperparameter is given explicitly as the Nishimori condition. As a natural extension of the conventional maximum likelihood method, our estimation $\hat{P}_N^\xi$ is realized by the average of probability models with Boltzmann weights. The result $\hat{P}_N^\xi$ is the best estimation obtained from the sample $\{\xi_i\}_{i=1, \cdots, N}$ as shown in Section~\ref{sec_limit}. The performance of the estimations $\hat{P}_N^\xi$ is determined by the set of prepared models $\mathcal{M}$. Therefore, it is necessary to find $\mathcal{M}$ that provides high estimation accuracy. In Section~\ref{sec_imp}, we have discussed the method of sequential extension to obtain such a set. In addition, we have discussed two interpretations of our method in Section~\ref{sec_interp} and simple examples in Section~\ref{sec_example}. 

Our method is valid when the data $\{\xi_i\}_{i=1, \cdots, N}$ are given independently. 
On the other hand, it cannot be applied directly to a sequence of correlated data encountered in many practical situation. 
In the sense of prediction, there is no use in estimating the distribution $P_{\rm GT}(\xi_i)$ for a single data $\xi_i$ when $\{\xi_i\}_{i=1, \cdots, N}$ are correlated. 
For a sequence of correlated data, it would be more meaningful to estimate the distribution $P_{\rm GT}(\xi_1, \cdots, \xi_N)$ of the series of the entire data $\{\xi_i\}_{i=1, \cdots, N}$. 
In this case, we should focus on estimating the mechanism that generates the data series. 
For example, for the series of data $\{\xi_i\}_{i=1, \cdots, N}$ generated by a stationary Markov process, estimating the transition probability $T(\xi_{i+1}|\xi_i)$ is more fondamental than estimating the distribution $P_{\rm GT}(\xi_1, \cdots, \xi_N)$. 
In this case, $\xi_{i+1}$ can be considered to be generated independently with given state $\xi_i$. 
Thus our method can be applied to estimate the transition probability $T(\xi_{i+1}|\xi_i)$. 

Through our proposed method, the average of multiple models weighted by KL divergence gives a better model. 
However, it is impractical to calculate $\hat{P}_N^\xi$ concretely in general because a partition function differs for each model. 
For example, even if the models are limited to the Boltzmann machines~\cite{ackley1985learning, hinton2012practical}, it is not easy to calculate a partition function for each model. 
On the other hand, such an average of multiple models is often used in ensemble learning~\cite{zhang2012ensemble}. 
Therefore, the difficulties faced in model averaging are the same as those encountered in ensemble learning. 
Our method is practically applicable for a set of probability models that are handled in ensemble learning. 
Ensemble learning tries to overcome them with various ideas. 
For example, parallel computation is effective since each model can be built independently. 
Random forests take a simple arithmetic mean of the expected values from many models, but from our point of view, we conclude that an average weighted by KL divergence is more reliable. 
Our proposed method is expected to help improve the accuracy of ensemble learning. 

The gauge symmetry of the KL divergence plays a central role in the maximum likelihood estimation. In supervised machine learning with a loss function, it would be possible to perform a regularization similar to our method by exploiting the gauge symmetry of the loss function. In the training phase of conventional machine learning with many parameters, such as deep learning, the model that minimizes the loss function is solely selected. In such a case, any loss function (and any regularization) gives almost the same results as long as the model is not significantly different from the ground truth. A model with high generalization performance is obtained empirically with little dependence on the form of the loss function (and regularization term) in deep learning with sufficiently large $\mathcal{M}$~\cite{zhang2021understanding, zhang2017musings}. However, in our framework, the generalization performance is expected to be highly dependent on the form of the loss function since models with high loss also contribute to the results. It is future work to understand the relationship between the choice of the loss function and generalization performance in machine learning from the viewpoint of gauge symmetry. 

\bibliography{MLM.bib}

\end{document}